\documentclass[a4paper,12pt]{article}
\usepackage{graphicx}
\begin{document}

\title{AI in arbitrary world}
\author{Dimiter Dobrev\\
Institute of Mathematics and Informatics\\
Bulgarian Academy of Sciences\\
Sofia 1113, BULGARIA\\
e-mail: d@dobrev.com}
\renewcommand{\today}{April 14, 2005}

\maketitle

\begin{abstract}
In order to build AI we have to create a program which copes well in an arbitrary world. In this paper we will restrict our attention on one concrete world, which represents the game Tick-Tack-Toe. This world is a very simple one but it is sufficiently complicated for our task because most people cannot manage with it. The main difficulty in this world is that the player cannot see the entire internal state of the world so he has to build a model in order to understand the world. The model which we will offer will consist of final automata and first order formulas.
 \end{abstract}

\section{The world of the Tick-Tack-Toe game}

In this paper we will observe a concrete artificial world. You can find this world and test it in the examples of the compiler Strawberry Prolog \cite{D1}. You have to start the example program {\bf World 2.spj}. After that you will see one panel with five lamps, three checkboxes and one button. You can observe this program as a test for intelligence. Really, it is made to distinguish AI but you can use it also to test a human being for its level of intelligence.

Here are the rules of the test. You are a step device which lives in an artificial world. To make a step you have to select your move in the checkboxes and to press the button ``Next Step''. What is your purpose? In order to cope well in this world you have to make more victories, and less losses and bad moves. You make victory when the lamp named ``Victory'' flashes. Respectively you make loss or bad move when the respective lamp flashes.

\begin{figure}
\begin{center}
\includegraphics[height=40mm]{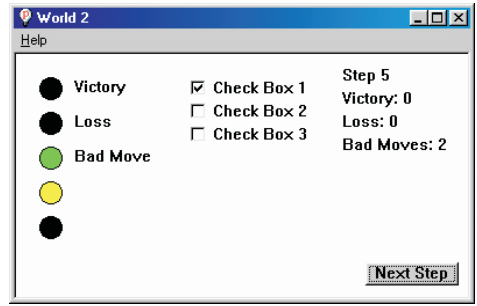}
\end{center}
  \caption{The panel}
\end{figure}

Of course, to cope well in this world you have to understand it first. This is extremely difficult even if you know that behind the panel is hidden the game Tick-Tack-Toe. The main problem is that you do not see the entire internal state of the game. This is obvious because you see only five lamps which do not carry enough information. Anyway, if you spare enough time for experiments in this world then you will understand it and will start to cope well in it. I am sure that you can manage in this artificial world because you manage to cope well in the real world, which is much more complicated. Really, you spare about 20 years for education before understanding the real world.

If you try to understand $World 2$ alone then you will understand the difficulties, which AI program has to overcome. It is better to try first the $World 1$. It is much easier because there is no hidden information in it. On every step you can observe the internal state of the $World 1$. In other words, the function $View$ which gives the lamps status from the internal state of the world is injective.

\section{A definition of AI}

We are going to make a program which is able to cope in $World 2$ because we want to build AI. Our first question is what is AI. We will accept the definition for AI which is given in \cite{D2,D3,D4,D5}. Here is a short description of this definition.

For us AI will be a step device which copes well in an arbitrary world. What is a world? For us this is a set $S$ of internal states, one starting state $s_{0}$ and two functions $World(s, d)$ and $View(s)$. The function $World$ will return the new state of the world from the current state and from the device's move (the state of the checkboxes). The function $View$ will inform us what does our device see. That is, this function will return the state of the lamps from the current state of the world. One device copes better than another if it makes more victories and less losses.

This definition differs from the so called Turing's test \cite{T1,T2,T3} because it separates the intellect from the education. Alan Turing described a device which copes well in the natural world. Actually, even the human being cannot do this without education. So his device is already educated before the start of the test or his device is specially built for the natural world and includes within itself all information about the real world. In the case with $World 2$ it is not a problem to build a program which copes well in Tick-Tack-Toe world. There are many programs which play this game very successfully (one of them you can find in [1]). Our goal is not to build a program specialised for $World 2$ but for a class of worlds as big as possible. The best case is if the real world is inside this class.

So we will suppose that our program does not know anything for the world before it starts living (making steps). The AI program has to collect all information alone. This will make the task of building of AI program extremely difficult.

\section{Detailed description of World 2}

In the $World 2$ you play Tick-Tack-Toe against an artificial partner which we will call Tom. The Tom's strategy is very simple. He plays every time randomly a correct move. You cannot see the whole board of the game but at every step you can see one cell by the state of the two yellow lamps. If the first lamp is on then this mean a cross. If the second lamp is on then this mean ``O''. If no lamp is on then there is nothing in the cell. Both lamps are never on together.

With the checkboxes you can select eight possible commands (moves). Four of them move your eye through the board. One command puts a cross at the position of the eye. Another one is the new game command. The last two commands are not used and they give bad move every time when you play them. When one Tick-Tack-Toe set is finished then one of the lamps Victory or Loss flashes depending on who won the set. If the set is equal then both lamps flash simultaneously. When after the end of the set you play the new game command then all cells are cleaned and become empty. The $World 2$ is made in this way because after the end of the set you will be able to continue to observe the Tick-Tack-Toe board in order to understand why you lost or won. Such observation is important for understanding of this world.

Every time when you try to do something which is not allowed the lamp ``Bad Move'' will flash red and the internal state of the world will stay unchanged.

Examples of bad moves: When you are in the first column and try to move left. When you try to put cross in the cell which is not empty or when the set is over. When you give the command new game before the end of the set.

\begin{figure}
\begin{center}
\includegraphics[height=40mm]{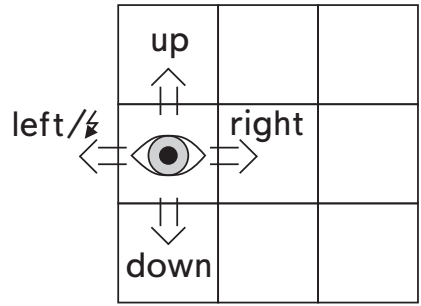}
\end{center}
  \caption{The World 2}
\end{figure}

\section{Model of the World 2}

In order for AI to understand the $World 2$ it has to build a model of this world. Here we will build such a model, which will consist from very simple parts. Most of these parts will be final automata with maximum three states. For such small automata exists possibility AI to find them by brutal force or by more creative approach. For bigger automata there is not such possibility due to the combinatory explosion.

The automate (1) which will describe the position of the eye is the following one:

\begin{figure}
\begin{center}
\includegraphics[height=25mm]{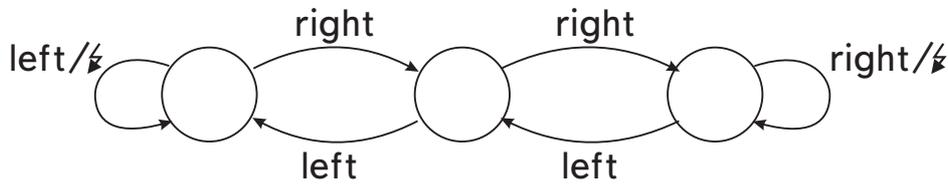}
\end{center}
  \caption{Automate 1}
\end{figure}

There are three states in this automate and they correspond to the left, middle and right column of the board. The change of the state occurs when AI makes the move ``left'' or ``right''. All other moves do not change the state of this automate. Here the flash stands for a ``bad move''. This means that if you try to play ``left'' from the left column then the lamp ``bad move'' will flash. Actually, this is the peculiarity of this automate, which will allow AI to find it. Actually, there is a huge number of final automata with three states but this one is special because it is connected with the rule that if the automate is in its first state and if AI plays ``left'' then the result is ``bad move''. From the other two states the move ``left'' every time is correct.

Of course, this automate is fundamental for the understanding of the $World 2$ because it gives to AI the information in which column the eye is. Analogicaly, AI will find the automate (2), which will say in which row the eye is and this two automata together will give the co-ordinates of the eye. Also analogicaly, AI will find the automate (3) with two states which will give information whether the game is over or not and when you can play the ``new game'' command. Interesting in this automate is that it can change its state due to the lamp status (if one or both of the lamps ``victory'' or ``loss'' flash).

\begin{figure}
\begin{center}
\includegraphics[height=30mm]{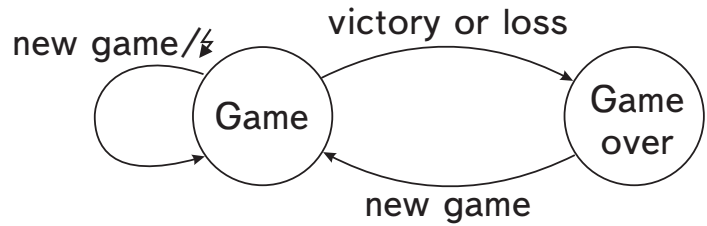}
\end{center}
  \caption{Automate 3}
\end{figure}

In order to begin to understand the rules and to play without ``bad moves'' we will need some easy rules like: If AI sees cross and plays ``put cross'' then the result is a ``bad move''. Simple rules like this can be represented like final automate with a single state.

In this way by constant rules and by deterministic automata we described the rules of $World 2$ and we can start playing correctly in it (without bad moves). Anyway this is not enough in order to understand this world and to cope well in it. For this purpose we need to have an idea for the situation on the board (for the game status). Unfortunately, all possible situations on the board are too much. They are 3 on the power of 9. Of course, not all of these situations are really possible but they are too much anyway and it is impossible for AI to find an automate with so many states by brutal force.

That is why we will describe the situation on the board in a different way. We will introduce a special undeterministic final automate which will be of second level because it will change its states depending on the condition of other automata. The first level automata, which we will use for this will be automata 1, 2 and 3. Actually, we will build 9 new automata - one for every possible co-ordinates $(X, Y)$ of the eye. The condition ``At X, Y'' will be true if the automate (1) is at the state $X$ and automate (2) is at the state $Y$. The condition ``Game over'' will be true if the automate (3) is at the state ``Game over''.

\begin{figure}
\begin{center}
\includegraphics[height=40mm]{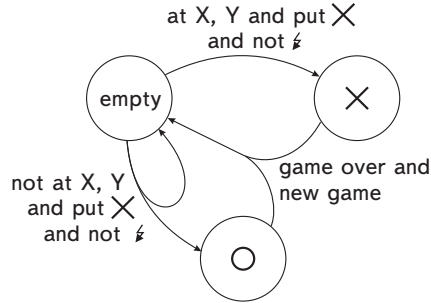}
\end{center}
  \caption{Automata (X, Y)}
\end{figure}

What is the peculiarity of this automate, which will allow to AI to find it? This is the fact that when the automate $(X, Y)$ is at the state ``cross'' and when the condition ``At X, Y'' is true then AI sees ``cross''. This peculiarity will also help to determine which state this undeterministic automate is in.

After this we can understand the $World 2$ and the fact that if AI puts cross then Tom will answer immediately by putting ``O'' on the board. Actually we need also the rule that Tom cannot put more than one ``O'' as a response. To describe such rule we need first order formula which can look something like this:

$$  \forall T \, \forall A,B (appear(O, A, T) \, \& \, appear(O, B, T) \rightarrow A=B)$$

This formula we can read as follows. ``For every moment $T$ if O appears in the cell $A$ and in the cell $B$ then $A$ and $B$ are the same cell.'' Here the universum of this formula is the set of moments (the natural numbers) plus the set of cells (the (X, Y) automata). We need one predicate $isO(A, T)$ which will be true if the automate $A$ is in state ``O'' at the moment $T$.  Then we can accept that $appear(O, A, T)$ is the formula $isO(A, T) \, \& \, not(isO(A, prev(T)))$. Here function $prev(T)$ gives the previous moment of the moment $T$.

With this formula AI will start to understand the game but it needs also to understand why it win or lose and for this it needs the concept of ``line'' and ``diagonal'' on the board. After this we will have the model of $World 2$ and we can start planing the future in this world. To plan its next move AI can use the algorithm Min-Max like the other programs, which play Tick-Tack-Toe.

This paper gives an idea of how we can make a program which will cope successfully in a Tick-Tack-Toe world without knowing apriory anything about this world. This program will be not that far from AI.


 \bibliographystyle{splncs}

\end{document}